\pdfoutput=1

\documentclass[11pt]{article}

\usepackage{emnlp2021}

\usepackage{times}
\usepackage{latexsym}

\usepackage[T1]{fontenc}

\usepackage[utf8]{inputenc}

\usepackage{microtype}

\usepackage{multirow}
\usepackage{hyperref}
\usepackage{cleveref}
\usepackage{url}
\usepackage{pgfplots}
\usepackage{booktabs}

\title{Alleviating Exposure Bias via Contrastive Learning for Abstractive Text Summarization}

\author{Shichao Sun \and Wenjie Li \\
  The Hong Kong Polytechnic University \\
  \texttt{csssun,cswjli@comp.polyu.edu.hk}
   \\}

\begin{document}
\maketitle

\begin{abstract}
Encoder-decoder models have achieved remarkable success in abstractive text summarization, which aims to compress one or more documents into a shorter version without the loss of the essential content. Unfortunately, these models mostly suffer a discrepancy between training and inference, i.e., the exposure bias problem. During the training stage, with \textit{teacher forcing} these models are optimized to maximize the likelihood of the gold summary given the gold summary tokens as input to the decoder, while at inference the given tokens are replaced by the generated tokens. Consequently, low-quality summaries are very likely to be generated. To remedy this problem, we propose to leverage \textit{contrastive learning} to decrease the likelihood of these low-quality summaries, and meanwhile increase the likelihood of the gold summary. Since our solution expands the states that the model perceives during training, we expect that the exposure bias problem can be alleviated. We experimentally demonstrate that our method effectively improves the performance of the state-of-the-art model on different datasets.\footnote{Our code is available at~\url{https://github.com/
ShichaoSun/ConAbsSum}}
\end{abstract}

\section{Introduction}
Abstractive text summarization is a task to create a short text according to one or several related documents, while at the same time preserving salient information. Recently, encoder-decoder models with pre-training such as BART~\cite{lewis2019bart} and PEGASUS~\cite{zhang2020pegasus} have achieved state-of-the-art (SOTA) performance in this task. The architectures of these models are commonly based on Transformer~\cite{vaswani2017attention} inspired by its great success in many Natural Language Generation (NLG) tasks. Essentially, their impressive achievements derive from pre-training. During this stage, models learn knowledge from massive unlabeled data through elaborate self-supervised objectives. After pre-training, these models are fine-tuned with \textit{teacher forcing} in downstream NLG tasks, like abstractive text summarization.

However, as a result of teacher forcing, these models mostly suffer a gap between training and inference~\cite{bengio2015scheduled}, like most of the vanilla sequence-to-sequence (seq2seq) models~\cite{sutskever2014sequence}. Their training objectives are to maximize the likelihood of each token in the gold summary given its previous tokens. At inference, however, the tokens of the gold summary are unavailable and they have to be replaced by the generated tokens.  It means that a model is often optimized under a limited part of the state space at the training time, so its performance is very likely to suffer degradation. This problem is well known as the exposure bias~\cite{ranzato2016sequence}. It may lead to the serious errors that accumulate quickly along the generated tokens. As a result, these models are peculiarly prone to producing the unexpected summary, which we term as ``silver summary'' in contrast to gold summary. What is worse, because of this problem, the silver summaries often contain fake facts~\cite{cao2018faithful,yuan2020fact}, which may look similar to the text in its surface form but actually contrary to its original meaning.

To alleviate the exposure bias, we propose to leverage the \textit{contrastive learning} method to expand the states that the model perceives during training. We expect not only to increase the likelihood of the gold summary via Maximum Likelihood Estimation (MLE) but also to decrease the likelihood of the silver summary via Contrastive Learning (CL) during the training process. It helps to prevent the model from generating the silver summary to a certain degree. This method can also be regarded as a special data augmentation strategy, which enables the model to learn from both the positive sample (gold summary) and the negative sample (silver summary). When the silver summary that is generated at inference explicitly participates in another round of training, the discrepancy between training and inference can reduce and in turn, the exposure bias can be mitigated.

We conduct experiments to verify our method on three benchmark datasets, including \textbf{Xsum}~\cite{narayan2018don}, \textbf{CNNDM}~\cite{nallapati2016abstractive} and \textbf{Multi-News}~\cite{fabbri2019multi}. The experimental results show that our methods can effectively improve the performance of a lately released SOTA model called PEGASUS.

\section{Method}
\subsection{Problem Definition}
\label{sec:problem}
The existing abstractive text summarization models mostly follow the seq2seq framework. They accept a long text as the input sequence and produce the output sequence as the corresponding summary . With teacher forcing, their learning objective is to maximize the likelihood of each token $y_{i}$ in the gold summary $Y$ with the conditions of the input text $X$ and the previous tokens $y_{<i}$ in the gold summary. The loss function is defined as negative log-likelihood (NLL) as follows: 
\begin{equation}
\label{eq:nll}
     L_{nll} = - \sum\nolimits_{i=1}^n f(y_i|X, y_{<i})
\end{equation}
where $f(y_i|X, y_{<i})$ is the log-likelihood of the $i$th token of the gold summary $Y$.

At inference, the models have to use the generated tokens $\widehat{y_{<i}}$ instead to predict the token $\widehat{y_i}$. Typically, the beam search algorithm is used to maintain multiple alternatives at each timestep based on the beam search score $S$. The models then produce the candidate summaries token by token via beam search and choose the one with the highest beam search score as the output summary. The beam search score of one alternative sequence $\widehat{Y}$ with $m$ tokens associated to the input text $X$ is calculated as follows:
\begin{equation}
\label{eq:inference_score}
    S(\widehat{Y}|X) = \frac{1}{m^\beta} \sum\nolimits_{i=1}^m f(\widehat{y_i}|X, \widehat{y_{<i}})
\end{equation}
where $f(\widehat{y_i}|X, \widehat{y_{<i}})$ is the predicted log-likelihood of the $i$th token of the generated sequence $\widehat{Y}$, $\widehat{y_{<i}}$ represents the tokens generated earlier than the token $\widehat{y_{i}}$, and $\beta$ is an additional exponential penalty associated to the sequence length. As for the task of text summarization, $\beta$ is smaller than 1.0 in order to avoid generating redundant information.

When training on the dataset via the NLL loss $L_{nll}$, the score $S$ of the gold summary is expected to rise so that it becomes more likely to produce the gold summary as one of the candidate summary and consequently chooses the gold summary as the final output among the generated candidate summaries.

However, there might exist a low-quality candidate that gets a higher score $S$. It is termed as ``silver summary'' when it is picked up as the output summary. The silver summary can be attributed to the discrepancy problem since the seq2seq model is only able to observe the gold summary at the time of training while the model needs to assess a large number of unseen alternatives at the time of inference. This problem is well-known as ``exposure bias''.

\subsection{Contrastive Learning}
\label{sec:con_loss}
To mitigate the above-mentioned problem, we propose to explicitly decrease the score $S$ of the silver summary during training via contrastive learning, which is inspired by the success of a similar contrastive learning method used in extractive summarization~\cite{zhong2020extractive}. 

Specifically, in our method, the seq2seq model is optimized to ensure that the ``pos score'' is higher than the ``neg score''. For the identical text $X$, the pos score $S(Y|X)$ is calculated via \Cref{eq:inference_score} using the gold summary, while the neg score $S(\widehat{Y}|X)$ is computed in the same way but using the silver summary. A margin ranking loss is defined to increase the pos score and meanwhile decrease the neg score as follows:
\begin{equation}
\label{eq:margin_loss}
    L_{con} = \max (0,S(\widehat{Y}|X) - S(Y|X) + \gamma)
\end{equation}
where $\gamma$ is a margin value.

Note that the model cannot be optimized if the pos score is higher than the neg score over the margin value $\gamma$, because when the value of $L_{con}$ is zero, the gradient is also zero. To make the most efficient use of training data and prevent the model from underfitting, we also include the NLL loss in the overall loss function, i.e.,
\begin{equation}
\label{eq:loss}
    L = L_{con} + L_{nll}
\end{equation}

\subsection{Model Training}

\begin{figure}[htbp] 
\centering 
\includegraphics[width=0.49\textwidth]{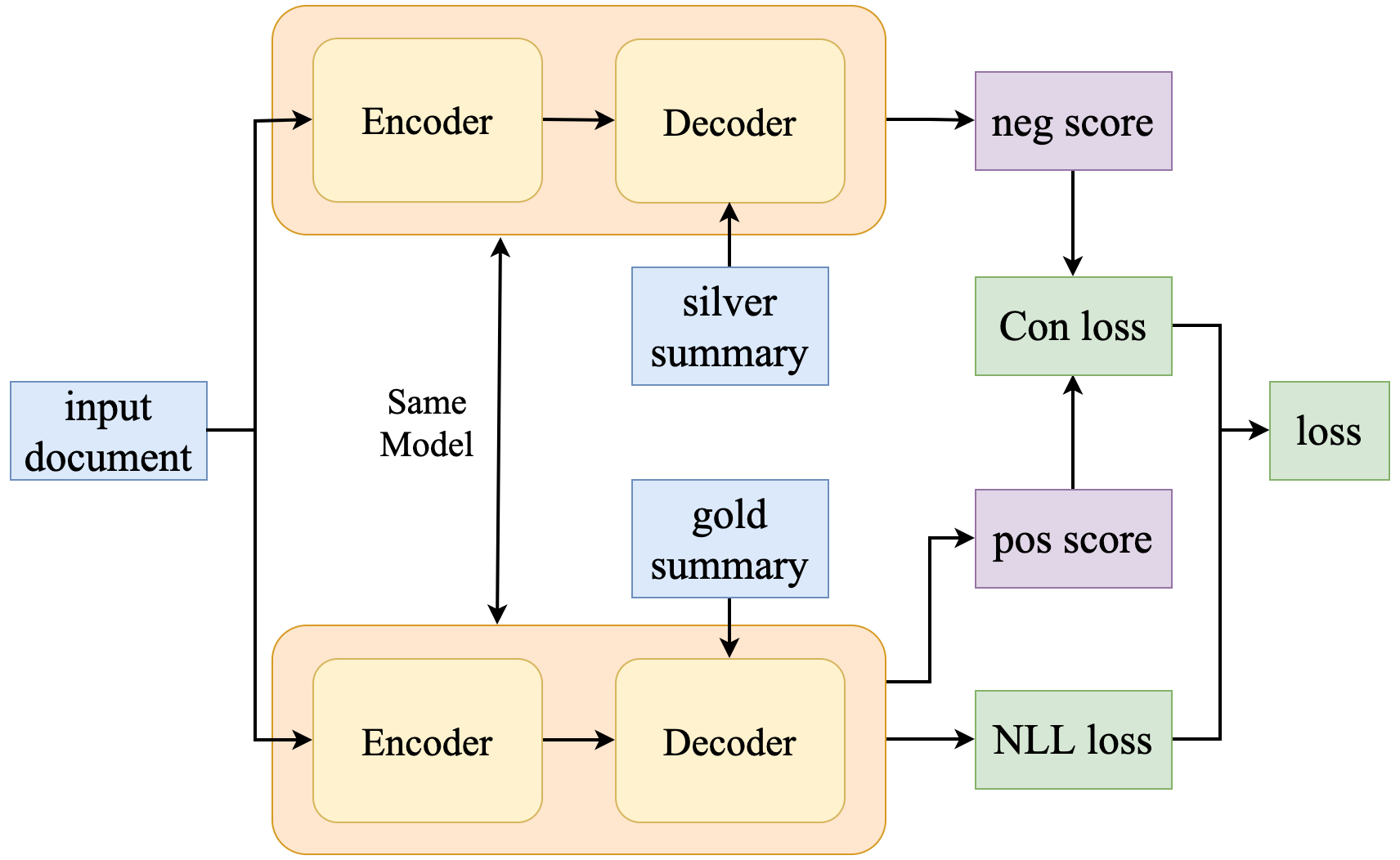} 
\caption{Model Training via Contrastive Learning.} 
\label{fg:framework}
\end{figure}

The working flow of training the seq2seq model via contrastive learning is illustrated in~\Cref{fg:framework}. Both the pos score and the neg score are computed are calculated based on the identical seq2seq model with the encoder-decoder architecture. Our method is not restricted to the vanilla seq2seq model, so the seq2seq model can also contain the copy mechanism and the coverage mechanism~\cite{see2017get}. While they share the encoder output for an input document, the two scores are respectively calculated by using the gold summary and the silver summary as the decoder input. It means that our method mainly imposes an effect on the decoder side. 

\section{Experiment}
We evaluate our method by testing whether it can improve the latest SOTA abstractive text summarization model called PEGASUS considering our method can be applied to almost any seq2seq model. We utilize the same encoder-decoder architecture as PEGASUS and initialize the parameters by the fine-tuned PEGASUS model. We freeze the encoder parameters to save computational resources and time as our method focuses on adjusting the decoder. We train the model via our contrastive learning method. 

Our code is implemented based on the Transformers~\footnote{\url{https://github.com/huggingface/transformers}} from Hugging Face, which well replicates PEGASUS model and provides the fine-tuned models for the evaluation purpose. More details of replication can be found in Appendix.

\subsection{Dataset}
We evaluate our method on three datasets, i.e., \textbf{XSum}, \textbf{CNNDM} and \textbf{Multi-News}, since they represent different kinds of text summarization tasks: (1)~\textbf{Xsum} contains news articles each associated with a summary. It is designed for single-document summarization and each summary contains one single sentence. (2)~\textbf{CNNDM} is a collection of news articles accompanied with several highlights as their summaries. It is also used for single-document summarization but the summaries often contain more than one sentence. (3)~\textbf{Multi-News} is a multi-document summarization dataset, which consists of news articles and human-written summaries. In general, Multi-News summaries are longer than CNNDM summaries.

\subsection{Baselines} 
To comparatively evaluate our method, we choose four SOTA models and one baseline method. \textbf{BERTS\small{UM}A\small{BS}}~\cite{liu2019text} enhances sentence-level BERT~\cite{devlin2018bert} to be the document-level encoder. \textbf{MASS}~\cite{song2019mass}, \textbf{BART}~\cite{lewis2019bart} and \textbf{PEGASUS}~\cite{zhang2020pegasus} are pre-trained models based on Transformers and they have obtained impressive performance. \textbf{Hi-MAP} \cite{fabbri2019multi} incorporates MMR into a pointer-generator network for multi-document summarization. \textbf{Scheduled Sampling}~\cite{bengio2015scheduled} (abbreviated to SS) is a conventional method to alleviate the exposure bias problem. We implement its two variants as baselines, where the decoder input (i.e., gold summary) is randomly replaced by the silver summary at either the summary level (denoted as sum) or the token level (denoted as token) with the probability of 0.5.

\begin{table*}[htbp]
\centering
\begin{tabular}{lcccccc}
\toprule
& \multicolumn{3}{c}{\textbf{Xsum}}  & \multicolumn{3}{c}{\textbf{CNNDM}} \\
& \multicolumn{1}{c}{Rouge-1} & \multicolumn{1}{c}{Rouge-2} & \multicolumn{1}{c}{Rouge-L} & \multicolumn{1}{c}{Rouge-1} & \multicolumn{1}{c}{Rouge-2} & \multicolumn{1}{c}{Rouge-L} \\
\hline
BERTS{\small{UM}}~\cite{liu2019text} & 38.76 & 16.33  & 31.15  & 41.72  & 19.39  & 38.76 \\
MASS~\cite{song2019mass} & 39.75 & 17.24  & 31.95  & 42.12  & 19.50  & 39.01 \\
BART~\cite{lewis2019bart} & 45.14  & 22.27    & 37.25  & 44.16  & 21.28 & 40.90 \\
\hline
PEGASUS (our implementation) & 47.08   & 24.53  & 39.28  & 44.17  & 21.44   & 41.26 \\
\quad + scheduled sampling (sum) & 47.25   & 24.63  & 39.35  & 44.32  & 21.40   & 41.39 \\
\quad + scheduled sampling (token) & 47.26   & 24.63  & 39.35  & 44.28 & 21.41 & 41.37 \\
ConSum (our method) & \textbf{47.34}  & \textbf{24.67}  & \textbf{39.40}  & \textbf{44.53}  & \textbf{21.54}  & \textbf{41.57} \\
\bottomrule
\end{tabular}
\caption{The results in single-document datasets (Xsum and CNNDM).}
\label{tab:sresults}
\end{table*}

\begin{table}[htbp]
\centering
\begin{tabular}{lccc}
\toprule
& R1 & R2 & RL \\
\hline
Hi-MAP & 43.47 & 14.89 & 17.41 \\
\hline
PEGASUS & 47.44 & \textbf{18.97} & 24.75 \\
\quad + SS (sum) & 47.66 & 18.94 & 24.84 \\
\quad + SS (token) & 47.67 & 18.95 & 24.85 \\
ConSum & \textbf{47.70} & 18.95 & \textbf{24.86} \\
\bottomrule
\end{tabular}
\caption{The results in Multi-News dataset.}
\label{tab:mresults}
\end{table}

\section{Results and Analysis}
\subsection{Automatic Evaluation}
We automatically evaluate summary quality using Rouge~\cite{lin2004rouge}~\footnote{\url{https://github.com/google-research/google-research/tree/master/rouge}}. The Rouge scores of single-document summarization and multi-document summarization are presented in~\Cref{tab:sresults,tab:mresults}, respectively. The first block shows the results reported in their original papers. The second block shows the results of our implementation. {\bf ConSum} is the abbreviation of our method, which stands for \textbf{Con}trastive \textbf{Sum}marization. There is a very small gap between the results of our implementation of PEGASUS and its reported results. This results from the fact that the beam search logic is different from the original implementation\footnote{One possible reason from the Transformers contributor (\url{https://github.com/huggingface/transformers/issues/6844})}. 

As shown in~\Cref{tab:sresults,tab:mresults}, our method can commonly obtain the rise of Rouge-1, Rouge-2 and Rouge-L on all datasets except a slight drop of Rouge-2 on Multi-News. Maybe this drop is because the summaries on Multi-News are longer. With the rise of Rouge-1, more information can be contained in summaries. Besides, our method also outperforms scheduled sampling on Xsum and CNNDM, while it obtains competitive results on Multi-News. It shows that our method is likely to obtain a bigger improvement than scheduled sampling. These results can reflect that our method is an effective way to boost PEGASUS.

\subsection{Ablation Study}
The ablation study is conducted on CNNDM to evaluate the contribution of the NLL loss and the contrastive loss (Con loss), and the results are given in ~\Cref{tab:ablation}. It tells that the NLL loss and the Con loss are complementary since we can achieve better results by using both of them. Besides, without the NLL loss, the model degrades a lot (0.33) but a suitable margin value ($\gamma=0.0$) can reduce the margin to 0.09. This is because the model optimizes more slowly and avoids overfitting when more document-summary pairs obtain the gradient 0 with the smaller margin value.

\begin{table}[htbp]
\centering
\begin{tabular}{lccc}
\toprule
& R1 & R2 & RL \\
\hline
PEGASUS & 44.17  & 21.44   & 41.26 \\
ConSum & \textbf{44.53}  & \textbf{21.54}  & \textbf{41.57} \\
\quad - Con loss & 44.39  & 21.49  & 41.43 \\
\quad - NLL loss &  &  &  \\
\quad \quad $\gamma=1.5$ & 44.20  & 21.33  & 41.26 \\
\quad \quad $\gamma=0.0$ & 44.44  & 21.49  & 41.49 \\
\bottomrule
\end{tabular}
\caption{Results for ablation study on CNNDM dataset.}
\label{tab:ablation}
\end{table}

\subsection{Results on Different Margin Value}
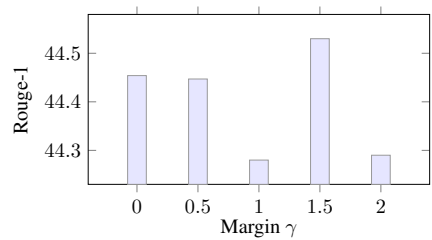
\begin{figure}[htbp]
\centering
\begin{tikzpicture}[scale=0.7]
\begin{axis}[ybar,enlargelimits=0.2,xlabel={Margin $\gamma$}, ylabel={Rouge-1},height=0.3\textwidth,width=0.5\textwidth] 
\addplot[draw=black!40,fill=blue!10] 
coordinates
{
 (0.0,44.454) (0.5,44.447) (1.0,44.28) (1.5,44.53) (2.0,44.29) 
};
\end{axis}
\end{tikzpicture}
\caption{Rouge-1 with different margin $\gamma$ on CNNDM}
\label{fig:margin}
\end{figure}

A suitable margin value can balance the NLL loss and the Con loss to obtain better performance. The performance is heavily influenced by the margin value, so we test the influence of margin $\gamma$ in the range \{0.0, 0.5, 1.0, 1.5, 2.0\}. The results in ~\Cref{fig:margin} show that our method can achieve the best result on CNNDM when $\gamma=1.5$. These results indicate that a small margin value is usually helpful to obtain competitive performance while the method is sensitive to the margin value.

\section{Conclusion}
In this paper, we propose to incorporate contrastive learning to train the abstractive summarization model as a means to alleviate the exposure bias resulting from teacher forcing. Experiments on three benchmark datasets indicate that our method improves the performance of the SOTA model and outperforms the scheduled sampling method. In the future, we would like to further explore whether this method can boost the performance of the SOTA models in other generation tasks such as question answer and response generation.

\bibliography{anthology,custom}
\bibliographystyle{acl_natbib}

\appendix

\section{Appendix}
\label{sec:appendix}
\subsection{Implementation Details}

We download the processed datasets from the websites~\footnote{\textbf{Xsum}:~\url{https://cdn-datasets.huggingface.co/summarization/xsum.tar.gz}; \\ \textbf{CNNDM}:~\url{https://cdn-datasets.huggingface.co/summarization/pegasus_data/cnn_dailymail.tar.gz}; \\ \textbf{Multi-News}:\url{https://cdn-datasets.huggingface.co/summarization/pegasus_data/multi_news.tar.gz}.}, where the three datasets have been well pre-processed for training and evaluation, especially for the PEGASUS model that we use as the baseline to evaluate our method.

The hyperparameters used for the different datasets are shown in~\Cref{tab:hparams}. The other hyperparameters are shared among all datasets. The dropout rate is 0.1 for all Transformer and attention layers. The learning rate is set to 1e-6 and weight-decay is set to 1e-8. We also use a learning rate schedule based on the cosine function along with linear warming-up. For fast validation, we only use 1000 random samples from the validation set instead of all validation data, and we validate the performance with a frequency of 0.01. The parameters of our model are optimized by Adafactor~\cite{shazeer2018adafactor}. We use an early-stopping strategy to obtain the best performance and reduce training time, where we monitor the Rouge-2 values and stop training when no improvement for 4 validation results is observed. 

\begin{table}[htbp]
\begin{tabular}{l c c c}
\toprule
Dataset & Xsum & CNNDM & M-News \\
\hline
batch size & 8 & 8 & 4 \\
length penalty  & 0.6 & 0.8 & 0.8 \\
doc length & 512 &  1024 & 1024 \\
sum length& 64 & 128 & 256 \\
margin value & 0.0 & 1.5 & 1.0 \\
\bottomrule
\end{tabular}
\caption{The hyperparameters related to datasets. M-News stands for Multi-News.}
\label{tab:hparams}
\end{table}
 
We conduct experiments on Xsum dataset by using 1 GTX1080Ti-12G GPU and spend at most 20 hours obtaining the model. And we train and evaluate my method on CNNDM and Multi-News datasets by using 1 RTX3090-24G GPU and spend no more than 30 hours obtaining the results.

\subsection{Replication}

We can replicate the experiments on the same computer, but our results become unstable when we use the different computer although we use the same hyperparameters. This is because our method is trained based on the generated silver summary, which can be altered as the rounding of different GPU varies. 

We usually need to adjust the hyperparameters to obtain a great improvement for different datasets or computer configurations. We empirically adjust the margin value or add scaling factor to NLL loss, while other hyperparameters mostly come from the PEGASUS fine-tuning configurations file\footnote{\url{https://huggingface.co/google/pegasus-large/blob/main/config.json}} and keep fixed. The margin value or scaling factor is usually set to a small value from 0.0 to 2.0. If the model overfits quickly, we usually reduce the learning rate with a decay rate 0.1 or 0.5, and simultaneously decrease the weight-decay using the same rate. In most cases, we can get a competitive result within 10 trials. 
\end{document}